\documentclass[10pt,twocolumn,letterpaper]{article}

\usepackage{wacv}
\usepackage[accsupp]{axessibility} 

\usepackage{graphicx}
\usepackage{amsmath}
\usepackage{amssymb}
\usepackage{booktabs}
\usepackage{mathtools}
\usepackage{adjustbox}
\usepackage{algorithm}
\usepackage{algpseudocode}

\usepackage[pagebackref,breaklinks,colorlinks]{hyperref}

\usepackage[capitalize]{cleveref}
\crefname{section}{Sec.}{Secs.}
\Crefname{section}{Section}{Sections}
\Crefname{table}{Table}{Tables}
\crefname{table}{Tab.}{Tabs.}

\DeclareMathOperator{\vspan}{span}
\DeclareMathOperator{\proj}{proj}

\begin{document}
\title{Approximating Intersections and Differences Between Linear Statistical Shape Models Using Markov Chain Monte Carlo}

\author{Maximilian Weiherer, Finn Klein, Bernhard Egger\\
Department of Computer Science\\
Friedrich-Alexander-Universtität Erlangen-Nürnberg\\
{\tt\small maximilian.weiherer@fau.de}}
\maketitle

\begin{abstract}
    To date, the comparison of Statistical Shape Models (SSMs) is often solely performance-based, carried out by means of simplistic metrics such as compactness, generalization, or specificity.
    Any similarities or differences between the actual shape spaces can neither be visualized nor quantified.
    In this paper, we present a new method to qualitatively compare two linear SSMs in dense correspondence by computing approximate intersection spaces and set-theoretic differences between the (hyper-ellipsoidal) allowable shape domains spanned by the models.
    To this end, we approximate the distribution of shapes lying in the intersection space using Markov chain Monte Carlo and subsequently apply Principal Component Analysis (PCA) to the posterior samples, eventually yielding a new SSM of the intersection space.
    We estimate differences between linear SSMs in a similar manner; here, however, the resulting spaces are no longer convex and we do not apply PCA but instead use the posterior samples for visualization.
    We showcase the proposed algorithm qualitatively by computing and analyzing intersection spaces and differences between publicly available face models, focusing on gender-specific male and female as well as identity and expression models. 
    Our quantitative evaluation based on SSMs built from synthetic and real-world data sets provides detailed evidence that the introduced method is able to recover ground-truth intersection spaces and differences accurately.
\end{abstract}

\section{Introduction}
\label{sec:intro}
Statistical Shape Models (SSMs) are a popular class of generative models providing a low-dimensional parametric representation of complex objects.
SSMs and especially 3D Morphable Models (3DMMs)~\cite{blanz1999morphable} are widely used within the computer vision community and often applied to model humans (faces, bodies, bones, and organs).
Their application ranges from face recognition~\cite{blanz2003face}, single-shot 3D face~\cite{feng2021learning, zielonka2022towards, li2021fit} and body reconstruction~\cite{loper2015smpl}, face reenactment~\cite{Thies15} and visual dubbing, to applications in the medical domain~\cite{luthi2017gaussian, fouefack2020dynamic, Weiherer23}, forensics, cognitive science, neuroscience, and psychology~\cite{Egger20}.
SSMs are typically built by applying Principal Component Analysis (PCA) to a set of objects in dense correspondence.
As such, an SSM is a \textit{linear} model in which shapes are represented as points in a low-dimensional, affine vector space.
Although other, non- or multi-linear models exist, PCA-based SSMs are still most common. 
They are easy to interpret, convenient to visualize, have excellent extrapolation capabilities, and are compatible with standard computer graphics pipelines.
Hence, we consider only PCA-based models in this work.
In general, however, the presented method can be used with all models that (i) use an affine vector space as an approximation for the underlying (manifold) shape space, and (ii) allow random sampling from and projecting shapes onto that space, including PCA-based Point Distribution Models~\cite{Cootes95} and Gaussian Process Morphable Models~\cite{luthi2017gaussian}, multi-linear~\cite{Bolkart16} and wavelet-based models~\cite{Brunton14}, non-linear models based on Principal Geodesic Analysis (by applying our method on the tangent space)~\cite{Fletcher04}, but also more modern approaches based on Variational Autoencoders~\cite{Kingma14}.

To date, comparison of SSMs is typically solely performance-based and carried out by applying metrics such as compactness, generalization, and specificity~\cite{davies2002learning, Styner03}.
A qualitative and direct comparison of the actual shape spaces spanned by two models is currently not possible; a visualization can only be provided for the individual shape spaces by inspecting random samples or the first principal modes of variation.
As such, any differences or similarities between two models' shape spaces can not be computed nor visualized with existing metrics.

In this paper, we present a new approach to qualitatively compare two linear SSMs that takes into account the affine vector spaces spanned by the models.
Specifically, we aim at computing the \textit{intersection} and \textit{set-theoretic difference} between the models' (hyper-ellipsoidal) allowable shape domains, \ie, those shapes that can either be explained by both SSMs or only by one model but not the other.
Given the extreme difference in dimensions between the low-dimensional shape spaces and their high-dimensional embedding, we propose to use a sampling-based technique to approximate the desired spaces.
Starting with a geometric motivation and formal definition, we formulate a probability distribution over shapes lying in the intersection or set-theoretic difference and use approximate inference based on Markov chain Monte Carlo (MCMC) to generate samples from those spaces.
Based on these samples, we compute the mean and a basis for the intersection space using PCA, eventually forming a new SSM that can be used for visualization, data exploration, and analysis.
Since the set-theoretic difference of two models' subspaces is no longer convex, we do not apply PCA but only visualize posterior samples.
Finally, note that thanks to the probabilistic nature of our method, we are not only able to inspect differences and similarities on shapes as a whole but also on a \textit{per-vertex} level, allowing us to understand which \textit{features} are present in both models and/or what is unique.

Our method has several interesting applications beyond a general data exploration use case.
For instance, differences between race-specific models (\eg, between Asian and White models) could allow for conclusions about demographic bias in SSMs.
In the medical domain, differences and similarities between two models built from a healthy and pathological group will help to visualize and identify novel phenotypes or shape-based clinical indicators.
Moreover, the difference between those two models or gender-specific shape variations may be of great interest as it could be later added to other models to increase variability, thus acting as a data augmentation strategy. 
Besides medical applications, we also see opportunities in product design.
When modeling body parts, the comparison of two SSMs from distinct populations may be helpful to improve the design for a specific market.
Finally, the difference between face identity and expression models would also reduce the effects of the identity-expression ambiguity in 3DMMs of faces~\cite{Egger21}, allowing for better expression neutralization or transfer. 

To summarize, the key contributions of this paper are: (i) we present a new method to qualitatively compare two SSMs by computing approximate intersection spaces and set-theoretic differences between the low-dimensional allowable shape domains spanned by the models, (ii) we provide an extensive, quantitative evaluation using models built from synthetic data sets and SSMs which are publicly available, (iii) we analyze intersections and differences between popular face models, including gender-specific male and female and identity and expression models, and (iv) we show how our method can be used to approximate intersection spaces and differences between texture models of 3DMMs.

\section{Related Work}
Although tackling a different problem, most related to our work is the method proposed by Hall et al.~\cite{Hall20}.
They present a \textit{splitting operation} used to remove one linear subspace from another and argue that this operation is seen as the inverse of the union operation.
It is important to understand that this method yields again a linear and convex subspace; they do \textit{not} compute the \textit{set-theoretic difference} between two linear models' shape spaces.
Rather, they split one subspace from the other, effectively keeping the intersection in that space. In contrast, the set-theoretic difference we are interested in excludes the intersection from this space (this is why we end up with a non-convex space). 
This allows us to investigate the \textit{true} difference between two models, \ie, those shapes that are exclusively represented in one model, which is not possible with~\cite{Hall20}.

\textbf{Performance-based metrics.}
The most well-known performance-based metrics to compare SSMs are compactness, generalization, and specificity~\cite{davies2002learning,Styner03}.
Those metrics allow to measure how well a model represents a certain population; they enable quantitative comparison of two models of the same population and are applied in most of the publications presenting novel SSMs.
However, although considered state-of-the-art when comparing SSMs, they do not allow for a visual inspection of similarities or differences between actual shape spaces.
Babalola et al.~\cite{Babalola06} proposed to use the Bhattacharya distance to measure the overlap between the two distributions implied by SSMs.
Although this might be useful in practice, it merely returns a scalar-valued distance.
Another common way to compare models is by evaluating their performance on downstream applications~\cite{booth20163d,gerig2018morphable,FLAME:SiggraphAsia2017,smith2020morphable}.
This, however, does not tell anything about what makes one model different or superior to another from a probabilistic point of view.
In general, none of the existing approaches enable visualization of similarities or differences between low-dimensional shape spaces.

\textbf{Computing intersections between vector spaces.}
Linear algebra offers an analytic way to compute the intersection between two subspaces embedded in a common vector space. 
Given bases $A,B\in\mathbb{R}^{m\times n}$ for two subspaces, a basis of the intersection space can be calculated by computing the null space of $C=[A,-B]\in\mathbb{R}^{m\times 2n}$.
When considering SSMs, however, we usually have $m\gg n$, \ie, there is an extreme difference in dimensions between the spanned subspaces of the models and their embedding.
In this case, the linear system that we would need to solve when computing the null space of $C$ is highly over-determined and may not have a non-trivial solution in general.

A lot of classical mathematical theory~\cite{anderson1969series,van1996matrix} is available to compute the orthogonal \textit{projection} matrix, $P$, used to project points onto the intersection of two subspaces.
Numerous formulas and methods have been proposed to compute $P$, of which the alternating projection method~\cite{von1949rings} is probably most well known.
However, this method has a high computational complexity and gives only the projection matrix instead of a basis for the intersection space.

Another method to calculate a basis for the intersection between two subspaces is by means of the Zassenhaus algorithm~\cite{luks1997some}.
More recently, Fenggang et al.~\cite{Fenggang19} proposed a closed-form algorithm to compute such a basis.

\section{Method}
After fixing notations and reviewing some basic properties of linear shape models, we first provide a formal definition of the intersection between two SSMs. 
We then present the algorithm to estimate the intersection. 
Additionally, we show how the same strategy can be further explored to estimate the differences between two models.

\subsection{Preliminaries}
\label{subsec:prelim}
A linear shape model can be interpreted either by assuming a linear algebraic or a probabilistic point of view.
Firstly, we take the linear-algebraic approach and represent an SSM as a \textit{function} of the form $f:\mathbb{R}^q\longrightarrow\mathbb{R}^{dn}$, $f(\alpha)=\bar{x}+U\alpha$, where $\bar{x}\in\mathbb{R}^{dn}$ is the so-called \textit{mean shape} computed over a set of objects described with $n$ points in $d$ dimensions (for triangular meshes, $d=3$).
The orthogonal matrix $U=(\sqrt{\lambda_1}u_1,\sqrt{\lambda_2}u_2,\dots,\sqrt{\lambda_q}u_q)\in\mathbb{R}^{dn\times q}$ holds the scaled eigenvectors of the dataset's covariance matrix, where $\lambda_i$ is the $i$-th eigenvalue.
The vector space spanned by a linear model is the affine subspace
\begin{equation}
    M\coloneqq\bar{x}+\vspan(U)=\{\bar{x}+U\alpha\mid\alpha\in\mathbb{R}^q\} \subseteq\mathbb{R}^{dn},
    \label{eq:model_affine_space}
\end{equation}
where $\dim(M)=q$.
Every linear combination of the basis vectors contained in $U$ gives rise to a new shape.

We now describe the probabilistic perspective on linear shape models.
By assuming $\alpha\sim\mathcal{N}(0,I)$, one can verify that shapes $x=f(\alpha)$ are distributed according to $x\sim\mathcal{N}(\bar{x}, C)$, where $C=UU^T\in\mathbb{R}^{dn\times dn}$ is the sample covariance matrix.
The probability of a shape is given by
\begin{equation}
    p(x)=p(\alpha)\propto\exp\left(-\frac{1}{2}\Vert\alpha\Vert_2^2\right).
    \label{eq:prob_shape}
\end{equation}
From this perspective, not every shape in $M$ is equally likely; the probability mass is concentrated in a $q$-dimensional hyper-ellipsoid which is centered at the mean $\bar{x}$ and whose principal axes correspond to the eigenvectors of $C$. 
The eigenvalues of $C$ are the reciprocals of the squares of the lengths of the semi-axes.
The shapes lying inside the hyper-ellipsoid can be characterized as $Q\coloneqq\{x\in M\mid (x-\bar{x})^TC^{-1}(x-\bar{x})\leq k\}\subseteq\mathbb{R}^{dn}$ with $k\in[0,1]$.
In terms of a probabilistic interpretation, this set can be equivalently rewritten as
\begin{equation}
    Q=\{x\in M\mid p(x)\geq\xi\},
    \label{eq:q_prob}
\end{equation}
where $\xi\in[0,1]$ (see supp. material for a derivation).
It contains all shapes with a probability greater than a certain threshold.
Following~\cite{Cootes04} and for a suitable $\xi$, shapes in $Q$ are considered \textit{plausible} in the sense that they look similar to the observed (training) data; hence, $Q$ is often called the \textit{allowable shape domain}~\cite{Cootes95}.

\subsection{Computing Intersections}
\label{subsec:comput_inter}
Given two aligned\footnote{By \textit{aligned}, we mean that Euclidean similarity transformations (rotation, translation, and scaling) between the two models have been removed. This can always be achieved by aligning the respective mean shapes using ordinary Procrustes analysis, see, \eg,~\cite{dryden16}.} SSMs in dense correspondence, our goal is to compute their intersection.
We define the intersection $I\subseteq\Omega=M_1\cup M_2$ between two SSMs as the intersection between their allowable shape domains, \ie, 
\begin{equation}
    I\coloneqq Q_1\cap Q_2.
    \label{eq:def_inter}
\end{equation}
In this work, we consider only SSMs with non-empty $I$ (whether two models intersect can be checked using, \eg,~\cite{Gilitschenski12}).
As seen from (\ref{eq:def_inter}) and by noting that $I=(M_1\cap M_2)\cap(Q_1\cap Q_2)$, a shape $x\in\Omega$ belongs to the intersection if the following conditions are met: \textbf{First}, $x$ can be \textit{represented} by both models, \ie, $x\in M_1\cap M_2$.
Assuming $x\in M_1\subset\Omega$, this is equivalent to finding an $x'\in M_2$ such that the (Euclidean) distance $d(x',x)$ vanishes (or vice versa if we assume $x\in M_2$).
\textbf{Second}, $x$ is \textit{likely} in both models, that is, $p(x)\geq\xi_1$ and $p(x')\geq\xi_2$.

Our definition is motivated by the fact that most of the shapes in the subspaces $M_1$ and $M_2$ are unlikely and do not lead to realistic shape instances.
Hence, if we defined the intersection simply as $I=M_1\cap M_2$, then $I$ would contain a lot of degenerated shapes that we want to disregard.
We therefore only consider the plausible regions of two models, which are exactly identified by $Q_1$ and $Q_2$.

The problem with the definition in (\ref{eq:def_inter}), however, is that real-world models are usually \textit{not} noise-free.
As a consequence, the first condition, $x\in M_1\cap M_2$, will never be met in practice since we can not find an $x'\in M_2$ such that $d(x',x)=0$ exactly.
To account for this, we weaken the strong definition and formulate an \textit{approximate} intersection $I_\epsilon$ for $\epsilon>0$ as
\begin{equation}
    I_\epsilon\coloneqq\left(Q_1\cap(Q_2+B_\epsilon(0))\right)\cup\left(Q_2\cap(Q_1+B_\epsilon(0))\right),
    \label{eq:approx_inter}
\end{equation}
where $B_\epsilon(0)=\{y\in\mathbb{R}^{dn}:d(0,y)<\epsilon\}$ is the $\epsilon$-ball in $\mathbb{R}^{dn}$ centered at $0$.
This allows us to also consider shapes to be in the intersection that are \textit{almost} (up to $\epsilon$) contained in $I$.
We can rewrite (\ref{eq:approx_inter}) as
\begin{equation}
    \begin{split}
        I_\epsilon=&\left\{x\in Q_1\mid\exists x'\in Q_2:d(x',x)<\epsilon\right\}\\
        \cup&\left\{x\in Q_2\mid\exists x'\in Q_1:d(x',x)<\epsilon\right\}.
    \end{split}
    \label{eq:def_approx_inter}
\end{equation}
Of course, $I_\epsilon\rightarrow I$ if $\epsilon\rightarrow 0$ and $I\subset I_\epsilon$ for every $\epsilon>0$.

\subsection{Algorithm}
Instead of explicitly constructing the set $I_\epsilon$, we aim at estimating the \textit{distribution} of points $x\in I_\epsilon$.
Given such distribution, we can then effectively generate samples from the intersection, which we use to build an SSM for the intersection by applying PCA.
The resulting linear model can then be utilized to visualize and study the intersection space.

The advantage of estimating the distribution of $x\in I_\epsilon$ instead of randomly sampling from $\Omega$ and applying hard constraints to test whether a point $x$ belongs to the intersection is that we do \textit{not} need to explicitly set values for the parameters ($\epsilon$, $\xi_1$, and $\xi_2$) involved in (\ref{eq:def_approx_inter}).

In detail, we model the posterior distribution of all $x\in Q_1$, given that $x\in I_\epsilon$ (and vice versa for all $x\in Q_2$):
\begin{equation}
    p(x \, \vert \, x\in I_\epsilon)=\frac{p(x\in I_\epsilon \, \vert \, x)p(x)}{\int p(x\in I_\epsilon \, \vert \, x)p(x)\text{d}x}.
    \label{eq:post}
\end{equation}
We call $p(x\in I_\epsilon \, \vert \, x)$ the likelihood function and denote it by $L(x;x\in I_\epsilon)$.
The probability $p(x)$ is computed according to (\ref{eq:prob_shape}) by noting that $x=f_1(\alpha)$ for $\alpha\in\mathbb{R}^{q_1}$.

The likelihood function encodes the conditions under which $x$ belongs to the intersection space.
In the following, we show how the two conditions as stated in Section \ref{subsec:comput_inter} can be directly translated into a likelihood.

\textbf{First condition.}
We implement the first condition, \ie, finding an $x'\in M_2$ such that $d(x',x)$ becomes small, using the orthogonal projection of $x$ onto $M_2$, given by
\begin{equation}
   x'=\proj_{M_2}(x)=f_2(\underbrace{U_2^{-1}(x-\bar{x}_2)}_{=\alpha'\in\mathbb{R}^{q_2}}).
   \label{eq:proj_operator}
\end{equation}
The point $x'$ is the closest point to $x$ contained in $M_2$ and hence minimizes the distance between $x$ and $x'$.
With that, we define a distance likelihood $L_D$ as
\begin{equation}
    \begin{split}
        L_D(x;x\in I_\epsilon)
        &\propto\exp\left(-\frac{1}{2}\left(\frac{d(x',x)}{\sigma}\right)^2\right),
    \end{split}
    \label{eq:dist_likelihood}
\end{equation}
where we used
\begin{equation}
    d(x',x)=\frac{1}{n}\sum_{i=1}^n\Vert x'_i-x_i\Vert_2
    \label{eq:def_mse}
\end{equation}
and assumed $d(x',x)\sim\mathcal{N}(0, \sigma^2)$.
A similar likelihood was originally proposed in \cite{Schoenborn17} and recently used by \cite{Madsen20} for model-based surface registration.

\textbf{Second condition.}
For the second condition, we require both $x$ and $x'$ to be likely.
This can be implemented by taking into account the probabilities $p(x)$ and $p(x')$ of $x$ and $x'$.
Since the probability of $x$ is already included in the formulation of the posterior in (\ref{eq:post}), we only need to care about $x'$.
To enforce the projection $x'$ to be likely, too, we define a projection likelihood $L_P$ simply as
\begin{equation}
   L_P(x;x\in I_\epsilon)=p(x'),
\end{equation}
where the probability of $x'=f_2(\alpha')$ is given by (\ref{eq:prob_shape}).
Our final likelihood is then a combination of the distance and projection likelihood and formulated as
\begin{equation}
   L=L_D L_P.
   \label{eq:combined_likelihood}
\end{equation}
The posterior distribution $p(x \,|\,x\in I_\epsilon)$ in (\ref{eq:post}) is now fully specified.
However, it is intractable and can not be computed exactly. 
We therefore make use of approximate inference as described in the following.

\subsection{Implementation}
We use Markov chain Monte Carlo (MCMC) to approximate the intractable posterior distribution in (\ref{eq:post}).
This is possible since $p(x \, | \, x\in I_\epsilon)\propto L(x;x\in I_\epsilon)p(x)$.

To construct a Markov chain, we make use of the Metroplis-Hastings (MH) algorithm.
The MH algorithm requires a proposal distribution conditioned on the current state.
We use the same random walk mixture proposal as proposed in \cite{Schoenborn17}.
Please refer to the supp. material for more information about the MH algorithm and our proposal distribution.
To explore the parameter space and to reduce auto-correlation between samples as much as possible, we use an ensemble of Markov chains instead of only one chain.
Each chain has a different starting point which is sampled from $\mathcal{N}(0,I)$.
We ensured that samples are sufficiently far apart to avoid individual chains exploring the same part of the parameter space.

\begin{figure*}
    \centering
    \includegraphics{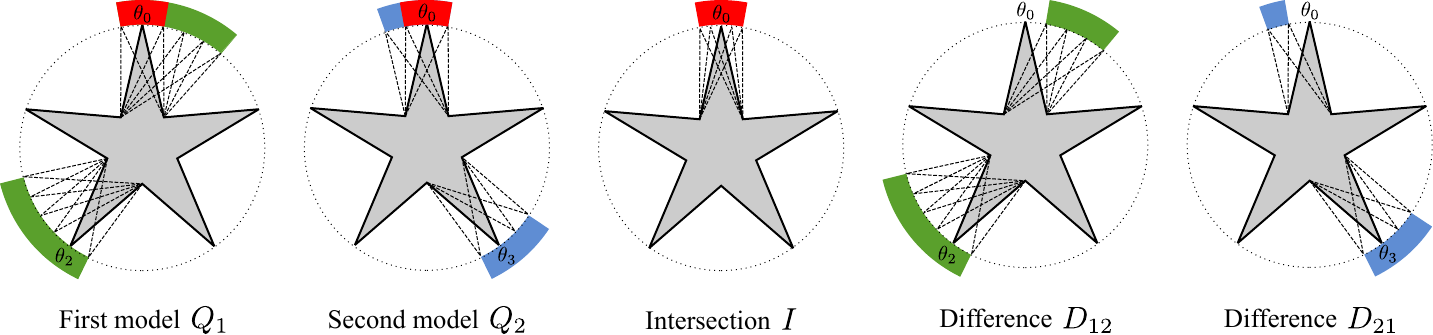}
    \caption{Schematic illustration of how we generate training data for star models including ground-truth intersections and differences. For the first model, we vary the first and third point of the star in $[\theta_j-a, \theta_j+b]$, where $j\in\{0,2\}$. For the second model, we vary the first and the fourth point in $[\theta_j-c, \theta_j+a]$, where $j\in\{0,3\}$. The ground-truth intersection contains stars where the first point varies from $[\theta_0-a, \theta_0+a]$, whereas the ground-truth difference of the first and the second model contains stars where the first point ranges from $(\theta_0+a, \theta_0+b]$ and the third between $[\theta_2-a, \theta_2+b]$.
    The ground-truth difference of the second and the first model includes stars where the first point varies from $[\theta_0-c, \theta_0-a)$ and the fourth from $[\theta_3-c, \theta_3+a]$.}
    \label{fig:star_data}
\end{figure*}

\subsection{Computing Differences}
Keeping in mind the definition of the approximate intersection space, we define the set-theoretic difference between two SSMs as
\begin{equation}
    D_{12}\coloneqq Q_1\setminus\left(M_2+B_\epsilon(0)\right).
    \label{eq:def_diff}
\end{equation}
The set $D_{12}$ contains all $x\in Q_1$ that can \textit{not} be represented in $M_2$ (or vice versa for $D_{21}$).
We can rewrite (\ref{eq:def_diff}) into
\begin{equation}
    D_{12}=\{x\in Q_1\mid\forall x^*\in M_2:d(x^*,x)\geq\epsilon\}.
    \label{eq:def_diff_all}
\end{equation}
From a computational perspective, this formulation is rather unpleasant due to the universal quantifier involved.
Note, however, that the point $x'=\proj_{M_2}(x)$ minimizes the distance from $x$ to $M_2$ and thus, if $d(x',x)\geq\epsilon$, then the same holds for all $x^*\in M_2$.
With this, we can rephrase (\ref{eq:def_diff_all}) as 
\begin{equation}
    D_{12}=\{x\in Q_1\mid d(x',x)\geq\epsilon\}.
    \label{eq:def_diff_proj}
\end{equation}
The condition of whether $x$ belongs to the difference can now be easily checked by means of the projection operator.

To estimate the distribution of points lying in $D_{12}$ (or $D_{21}$), we only have to make two small changes to the algorithm presented throughout the previous sections.
From the definition of the difference in (\ref{eq:def_diff_proj}) we observe that the only condition a point $x\in Q_1$ has to fulfill in order to lie in $D_{12}$ is that the distance to its projection onto $M_2$ becomes large.
This can be implemented by \textit{inverting} the distance likelihood in (\ref{eq:dist_likelihood}) as
\begin{equation}
    \bar{L}_D(x;x\in I_\epsilon)\propto 1- L_D(x;x\in I_\epsilon).
\end{equation}
Secondly, since we no longer require the projection to be likely, we remove the projection likelihood from (\ref{eq:combined_likelihood}).

After those minor modifications, our algorithm is ready to be used to approximate the distribution of shapes lying in the difference between two models.
However, the space in which the shapes of the difference between two SSMs lie is no longer convex as the shapes lying in the intersection have been removed.
As such, it does not make any sense to apply PCA to the MCMC samples as the resulting linearized space would simply interpolate the intersection.
Instead, we only visualize random samples drawn from our Markov chain in order to inspect the difference.

\section{Experiments and Results}
We conducted several quantitative and qualitative experiments to validate our method.

\begin{table*}
    \centering
    \begin{adjustbox}{width=1\textwidth}
        \begin{tabular}{ccccc|c||ccc|cc||cc}
            \toprule
            \multicolumn{5}{c|}{Training data} & \multicolumn{4}{c|}{\textbf{Intersection} (Grassmann distances, $d_G$ $\downarrow$)} &  \multicolumn{4}{c}{\textbf{Difference} (Reconstruction errors, $d_R$ $\uparrow$)}\\
            \cmidrule(lr){1-5} \cmidrule(lr){6-9} \cmidrule(lr){10-13}
            $Q_1$ & $Q_2$ & $I$ & $D_{12}$ & $D_{21}$ & $d_G(\hat{I},I)$ &  $d_G(Q_1,I)$ & $d_G(Q_2,I)$ & $d_G(Q_1,Q_2)$ & $d_R(\hat{D}_{12},I)$ & $d_R(\hat{D}_{21},I)$ & $d_R(D_{12},I)$ & $d_R(D_{21},I)$\\
            \midrule
            $[-5,40]$ & $[-20,5]$ & $[-5,5]$ & $(5,40]$ & $[-20,-5)$ & $0.0260\pm 0.0086$ & 1.5634 & 1.5634 & 2.1211 & $0.2235\pm 0.1132$ & $0.1148\pm 0.0534$ & $0.1781\pm 0.2243$ & $0.0911\pm 0.1122$\\
            $[-5,20]$ & $[-20,5]$ & $[-5,5]$ & $(5,20]$ & $[-20,-5)$ & $0.0370\pm 0.0110$ & 1.5583 & 1.5583 & 2.1211 & $0.1165\pm 0.0556$ & $0.1171\pm 0.0542$ & $0.0911\pm 0.1122$ & $0.0911\pm 0.1122$\\
            $[-10,40]$ & $[-20,10]$ & $[-10,10]$ & $(10,40]$ & $[-20,-10)$ & $0.0140\pm 0.0052$ & 1.5471 & 1.5506 & 2.1211 & $0.2081\pm 0.1097$ & $0.1150\pm 0.0540$ & $0.1867\pm 0.2110$ & $0.1073\pm 0.1081$\\
            $[-10,20]$ & $[-20,10]$ & $[-10,10]$ & $(10,20]$ & $[-20,-10)$ & $0.0323\pm 0.0094$ & 1.5506 & 1.5506 & 2.1211 & $0.1107\pm 0.0510$ & $0.1143\pm 0.0534$ & $0.1126\pm 0.1080$ & $0.1073\pm 0.1081$\\
            $[-20,60]$ & $[-30,20]$ & $[-20,20]$ & $(20,60]$ & $[-30,-20)$ & $0.0053\pm 0.0022$ & 1.5240 & 1.5440 & 2.1211 & $0.2692\pm 0.1571$ & $0.1509\pm 0.0809$ & $0.2834\pm 0.3170$ & $0.1861\pm 0.1569$\\
            $[-40,80]$ & $[-50,40]$ & $[-40,40]$ & $(40,80]$ & $[-50,-40)$ & $0.0018\pm 0.0004$ & 1.4863 & 1.5375 & 2.1211 & $0.3257\pm 0.1994$ & $0.2243\pm 0.1293$ & $0.4200\pm 0.3993$ & $0.3501\pm 0.2531$\\
            \bottomrule
        \end{tabular}
    \end{adjustbox}
    \caption{Quantitative results for the star models based on different training sets, averaged over five runs. The comparison of the estimated intersection space, $\hat{I}$, and the ground-truth intersection $I$ is based on the Grassmann distance, denoted as $d_G$. We also report various baseline distances to give an intuition for the range of the distances. To evaluate whether or not posterior samples from the computed differences, $\hat{D}_{12}$ and $\hat{D}_{21}$, are indeed from the true difference, we calculate reconstruction errors $d_R$ by projecting estimated samples and samples from the ground-truth differences $D_{12}$ and $D_{21}$ into the ground-truth intersection model $I$. Note that the intervals for the training data arise from setting different values for the parameters $a$, $b$, and $c$.}
    \label{tab:star_model}
\end{table*}

\subsection{Quantitative Analysis}
\label{subsec:quant_analysis}
Our quantitative evaluation is based on two data sets from which we generate ground-truth intersection spaces.
The first is a synthetic data set based on a five-pointed star.
Due to the rather simplistic geometry, this data set is well suited for an initial analysis and serves as a proof-of-concept.
The second is a real data set derived from a real-world SSM and, consequently, shares a lot of desired properties such as high dimensionality.

We employ the Grassmann distance as a natural distance between two affine subspaces to compare ground-truth intersections with the spaces recovered by our method.
The Grassmann distance is defined as the Euclidean norm of the affine principal angles between two subspaces and can be easily computed by applying Singular Value Decomposition to the Stiefel coordinates of the two spaces, see, \eg \cite{LekHeng20}.
Please refer to the supp. material for more information.

Constructing ground-truth differences is far more complex due to their topological structure.
As such, the quality of the differences recovered by our method can only be evaluated on the star models since, due to their synthetic construction, we are able to generate samples from the ground-truth difference (\ie, samples for which we definitely know they are lying in the difference).
This is not the case for real-world SSMs because \textit{the ground-truth difference space for real-world models is unknown}.

\subsubsection{Synthetic Data Set}
\label{subsubsec:synthetic}
For the synthetic data set, we use a five-pointed star as a simple geometric object to generate training data by systematically varying its points, see Figure \ref{fig:star_data}.
Each point of the star can be written as $(r\cos{\theta_i}, r\sin{\theta_i})$ with radius $r$ and angle $\theta_i$, $i=0,\dots,4$.
For the first model, we vary the first and third point in range $[\theta_j-a, \theta_j+b]$, where $j\in\{0,2\}$.
For the second model, we vary the first and fourth point in range $[\theta_j-c, \theta_j+a]$, where $j\in\{0,3\}$.
The ground-truth intersection model contains stars where the first point ranges from $[\theta_0-a, \theta_0+a]$.
The ground-truth difference of the first and the second model contains stars where the first point ranges from $(\theta_0+a, \theta_0+b]$ and the third between $[\theta_2-a, \theta_2+b]$.
The ground-truth difference of the second and the first model includes stars where the first point varies from $[\theta_0-c, \theta_0-a)$ and the fourth from $[\theta_3-c, \theta_3+a]$.

We tested our method on six different values for $a,b,c$ as shown in Table \ref{tab:star_model}.
An ensemble of 15 chains was used, where each chain was sampled 2,500 times.
1,000 samples were considered as burn-in.
Out of all samples, 5,000 samples were evenly chosen to build an SSM for the intersection space.
For the distance likelihood, we set $\sigma=0.003$ for all experiments (see supp. material for an ablation on $\sigma$).

We use the Grassmann distance as described above to measure how well our method can recover the ground-truth intersection space.
To evaluate the quality of the computed differences, we exploit the fact that we have \textit{samples} from the ground-truth difference model (clearly, we do \textit{not} have a parametrization of these models, but only a few samples thereof).
To this end, we quantify whether or not samples from the difference generated by our method have as high a distance from their reconstruction in the ground-truth intersection space as the samples from the ground-truth difference (note that a sample belongs to the difference if it can \textit{not} be represented in the intersection).
We, therefore, project the MCMC samples of the difference into the ground-truth intersection model and calculate the average reconstruction error, see supp. material for further details.
We do the same for the samples from the ground-truth difference.
Similar reconstruction errors then indicate that all samples generated by our method are indeed valid samples from the true difference space.

\begin{figure}
    \centering
    \includegraphics[width=\linewidth]{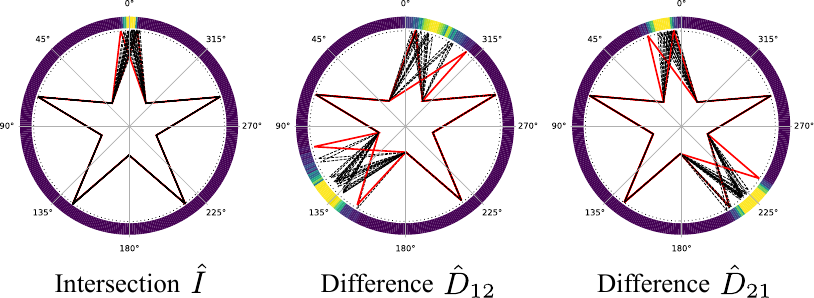}
    \caption{Visualization of (a random subset of) MCMC samples generated to compute the estimated intersection $\hat{I}$ and differences $\hat{D}_{12}$ and $\hat{D}_{21}$ between two star models. The included polar frequency histogram visualizes the distribution of posterior samples; it indicates that most of the samples are indeed from the true posterior distribution (\ie, they are within the bounds shown in red).}
    \label{fig:empirical_star}
\end{figure}

Overall, the proposed method is able to recover the ground-truth intersection space for all six models very well, see Table \ref{tab:star_model}.
We measure an average Grassmann distance of 0.0194 between ground-truth intersections and the intersection spaces recovered by our method.
We observe a similar result for the estimated differences in terms of reconstruction errors.
Figure \ref{fig:empirical_star} provides additional visualization of the distribution of posterior samples.

\begin{figure*}
    \centering
    \includegraphics[width=\linewidth]{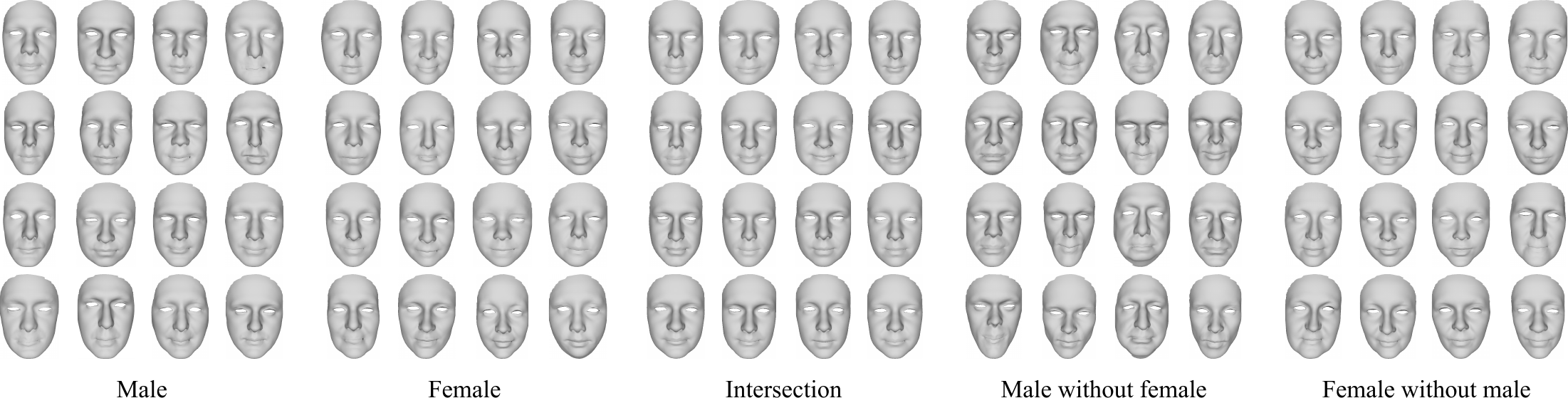}
    \caption{Random samples from the male and female model of the LYHM data set (1st and 2nd block) as well as samples drawn from the computed intersection model (3rd block) and the respective differences (male without female, and female without male, 4th and 5th block). We observe stronger male and female dominance in the differences and neutral gender in the intersection.}
    \label{fig:male_female}
\end{figure*}

\begin{table}
    \centering
    \begin{adjustbox}{width=1\linewidth}
        \begin{tabular}{ccc|c||ccc}
            \toprule
            \multicolumn{3}{c|}{Training data} & \multicolumn{4}{c}{\textbf{Intersection} (Grassmann distances, $d_G$ $\downarrow$)}\\
            \cmidrule(lr){1-3} \cmidrule(lr){4-7} 
            $\dim(Q_1)$ & $\dim(Q_2)$ & $\dim(I)$ & $d_G(\hat{I},I)$ & $d_G(Q_1,I)$ & $d_G(Q_2,I)$ & $d_G(Q_1,Q_2)$\\
            \midrule
            3 & 3 & 1 & $0.0032\pm 0.0010$ & 1.5517 & 1.5532 & 2.1854\\ 
            6 & 6 & 2 & $0.0645\pm 0.0455$ & 2.1641 & 2.1855 & 3.0835\\ 
            9 & 9 & 3 & $0.0877\pm 0.0178$ & 2.6375 & 2.6901 & 3.7747\\
            12 & 12 & 4 & $0.1683\pm 0.0106$ & 3.1004 & 3.0849 & 4.3563\\
            15 & 15 & 5 & $0.1939\pm 0.0433$ & 3.4606 & 3.4467 & 4.8682\\ 
            \bottomrule
        \end{tabular}
    \end{adjustbox}
    \caption{Quantitative results for the estimated intersection spaces of the real-world data set. We tested five different configurations of individual model dimensions, $\dim(Q_1)$ and $\dim(Q_2)$, and the dimension of the ground-truth intersection space, $\dim(I)$. Results were averaged over three different random splits per configuration.}
    \label{tab:bfm_model}
\end{table}

\subsubsection{Real Data Set}
The simple geometry and ground-truth intersection spaces of the star models enabled an intuitive analysis of the proposed method.
The geometry of real-world SSMs, however, is oftentimes far more complex. We conduct experiments on those models to further validate the proposed approach.

Given an SSM with a set of basis vectors, we generate two individual models and a ground-truth intersection by randomly splitting the set of basis vectors into three disjoint subsets, $S_1,S_2$, and $S_I$.
The first two sets, $S_1$ and $S_2$, contain the unique basis vectors for the first and second model, and the last set, $S_I$, holds the basis vectors of the ground-truth intersection.
The two models' bases are then given by $S_1\cup I$ and $S_2\cup I$, respectively.
Since we can not generate samples from the ground-truth difference, we present evaluations only for the computation of intersection spaces.

We used the first 15 basis vectors of the BFM 2019 \cite{gerig2018morphable} for this experiment.
Moreover, an ensemble of 25 chains was used.
Each chain was sampled 5,000 times with a burn-in phase of 2,000 samples.
To build the intersection model, again 5,000 samples were evenly chosen out of all samples.
We empirically set $\sigma=0.3$ for all models.

Table \ref{tab:bfm_model} presents the results for five different random splits.
As shown, the proposed method is able to recover all ground-truth intersection spaces quite well with an average Grassmann distance of 0.1035.
However, we observe increasing distances as the dimension of the intersection increases.
This is most likely because MCMC does not properly sample the entire space, and may be alleviated by choosing larger proposals or running more Markov chains.
Indeed, by doubling the number of chains, the Grassmann distance decreases by more than 10\%.

\subsection{Qualitative Results}
To showcase our method on real-world SSMs, we compute and analyze intersections and differences between (i) a gender-specific male and female model, and (ii) an identity and expression model.
Due to the missing ground-truth intersection spaces and differences, we can only provide qualitative results.
However, to underline that our method does \textit{not} simply produce the union of two models, or even worse, just reproduces one model, we do report Grassmann distances to the union and to individual models.
The union of two models is computed by applying PCA to a set of random samples from both models.

\textbf{Male and female models.}
Both models were built using 600 faces from the LYHM database \cite{Dai2019}.
We used a face mask to only include the frontal region of the face and truncated both models to include the first 50 basis vectors.
Each model has 5,764 vertices.
As such, both SSMs span a 50-dimensional affine subspace which is embedded in a 17,292-dimensional vector space.

Exemplary results can be found in Figure~\ref{fig:male_female}.
As seen, random faces sampled from the intersection model do not show a strong preference towards male or female, and most of the time tend to look more neutral compared to faces drawn from the individual male or female models.
Furthermore, while random faces from the difference between male and female show extremely masculine traits, samples from the difference between female and male look rather feminine.
We report a Grassmann distance of about 1.8 from the intersection to the union of male and female, and a distance of 3.2 and 2.8 to individual male and female subspaces.
Computing the intersection model took about 110 minutes on a single core; differences half the time.

\begin{figure*}
    \centering
    \includegraphics[width=\linewidth]{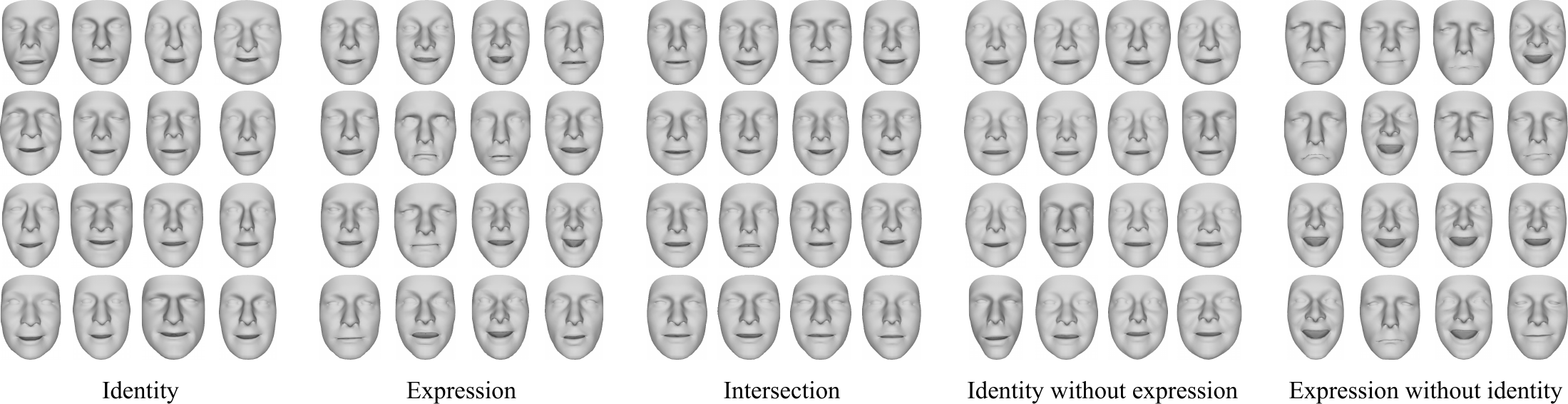}
    \caption{Random samples from the original identity and expression models of the BFM 2019 (1st and 2nd block) as well as samples of the computed intersection model (3rd block). Also shown are samples from the difference between identity and expression models (4th and 5th block). We observe less expression variation in samples drawn from the difference of identity and expression (identity without expression, 4th block) and less identity variation in the difference of expression and identity (expression without identity, 5th block).}
    \label{fig:id_expression_samples}
\end{figure*} 

\begin{figure}
    \centering
    \includegraphics[width=\linewidth]{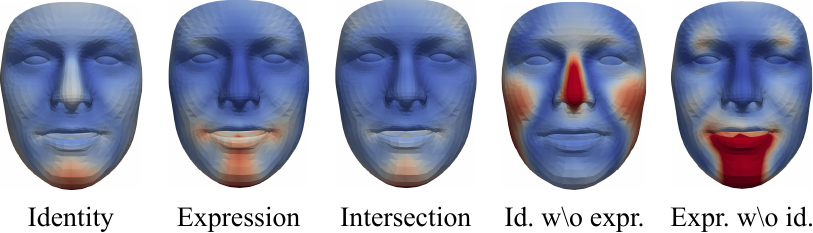}
    \caption{Per-vertex level visualization of similarities and differences between the BFM 2019 \cite{gerig2018morphable} identity and expression models. Depicted is the per-vertex variance of the original identity and expression models as well as the posterior variance for the computed intersection model and differences (red areas show high variance, blue areas low). The intersection visualizes the source of the identity-expression ambiguity, differences highlight \textit{features} represented either by identity (nose) or expression (opening of mouth, raising cheeks and eyebrows). 
    }
    \label{fig:identity_expression}
\end{figure}

\textbf{Identity and expression models.}
Face identity and expression are usually modeled separately in 3DMMs.
As recently demonstrated in \cite{Egger21}, however, the subspaces spanned by the identity and expression model are far from being orthogonal and thus, face identities and expressions are not independent of each other.
As a consequence, it is not possible to alter a specific face identity without leaving its expression \textit{completely} unchanged (and vice versa).
Although only hardly visible with the human eye, this effect has implications on downstream applications such as expression transfer and inverse rendering \cite{Egger21}.

We use our method to further study this ambiguity, also known as the identity-expression ambiguity, and analyze the intersection as well as differences between the identity and expression models of the BFM 2019 \cite{gerig2018morphable}.
We again truncated both SSMs to the first 50 basis vectors.
Each model has 1,746 vertices, leading to a 5,238-dimensional vector space in which the models' subspaces are embedded.

Results are shown in Figures~\ref{fig:id_expression_samples} and~\ref{fig:identity_expression}.
While there is no ground-truth for this task, the observations match our intuition, \eg, shape variations of the nose do not arise in the expression model but solely in the identity model, and effects based on muscle movement dominate in the identity model.
To still add a quantitative element, we measured the Grassmann distances between the intersection and original identity and expression models and report a distance of 5.36 to identity, and 5.09 to the BFM expression model.
Those distances being close to each other show that the intersection model somehow lies between the identity and expression model.
The Grassmann distance to the union is 3.48.
It took about 30 minutes to compute the intersection model. Again, computing differences required half the time.

For both tasks (male vs. female and identity vs. expression), we observed very similar results for the FLAME~\cite{Li17} model. Please refer to the supp. material.

\subsection{Extension to Color}
\label{subsec:extension_color}
Since texture in 3DMMs is modeled similarly to shape~\cite{blanz1999morphable}, our method can be used to also compute intersections and differences between texture models \textit{without} any algorithmic changes.
Results for male and female 3DMMs described above can be found in the supp. material.

\section{Limitations}
Although the presented method yields appealing results, our approach in its current form is not without limitations.
First, with increasing dimension of the intersection space more Markov chains are needed to explore the parameter space; this phenomenon is known as the curse of dimensionality.
It naturally increases the run time of our method.
We plan to tackle this issue by parallelizing Markov chains and by using more effective MCMC samplers such as the Metropolis-adjusted Langevin algorithm~\cite{Girolami11}.

Second, if SSMs are not in dense correspondence, we require some form of optimization in the projection operation, \ie, (\ref{eq:proj_operator}) needs to be adapted.
Clearly, while the projection can be computed in closed-form if models are in correspondence, we require registration if they are not in correspondence. 
This can be \textit{efficiently} carried out \textit{prior} to the application of our method using techniques proposed in~\cite{ploumpis2019combining}.

Lastly, although our method and the chosen parameters might not be trivial to evaluate on real-world models, the result can \textit{always} be well validated empirically by estimating $\epsilon$ from the MCMC samples, see supp. material.

\section{Conclusion}
In this paper, we have introduced a new method to compare SSMs by computing approximate intersection spaces and set-theoretic differences between the low-dimensional allowable shape domains spanned by two models in dense correspondence.
We showed how MCMC can be leveraged to approximate the distribution of shapes lying inside those spaces, and, based on posterior samples, compute a new SSM for the intersection.
Building upon this algorithm, we further demonstrated that our method can be easily adapted to also compute set-theoretic differences.
Confirmed by quantitative and qualitative evaluation on synthetic data and real-world face models, the proposed method is able to recover ground-truth intersection spaces and differences accurately.
Moreover, qualitative results obtained on real-world SSMs match our intuition.
In future work, we plan to study how our approach is (or may be) correlated to performance-based metrics, such as the Bhattacharya distance.

\vspace{0.15cm}
\footnotesize
\noindent\textbf{Acknowledgments.}
We thank Tinashe Mutsvangwa, Marcel Lüthi, and Tobias Meier for interesting discussions and feedback.
This work was funded by the German Federal Ministry of Education and Research (BMBF), FKZ: 01IS22082 (IRRW), and the FAU Emerging Talents Initiative (ETI).
The authors are responsible for the content of this publication.

{\small
\bibliographystyle{ieee_fullname}
\bibliography{PaperForArxiv}
}

\clearpage

\normalsize

\appendix
\section*{Appendix}
In this supplementary material, we provide (i) additional information on the equivalence of the two definitions for $Q$ in Section \ref{subsec:prelim} of the main paper, (ii) further implementation details on our algorithm, (iii) additional information on the evaluation metrics used in Section \ref{subsec:quant_analysis} of the main text, (iv) an ablation study on the key parameter of our method, $\sigma$, (v) a practical strategy on how to empirically choose $\sigma$, and finally (vi) some additional qualitative results.

\section{Equivalence of Q}
In Section \ref{subsec:prelim} of the main paper, the set $Q$ was defined as 
\begin{equation}
    Q\coloneqq\{x\in M\mid (x-\bar{x})^TC^{-1}(x-\bar{x})\leq k\}\subseteq\mathbb{R}^{dn},
\end{equation}
where $k\in[0,1]$.
We claimed that, in terms of a probabilistic interpretation, $Q$ can be equivalently rewritten as 
\begin{equation}
    Q=\{x\in M\mid p(x)\geq\xi\}
\end{equation}
with $\xi\in[0,1]$, see Eq. (\ref{eq:q_prob}) of the main paper.
This can be easily verified by
\begin{equation}
    \begin{aligned}
        && (x-\bar{x})^TC^{-1}(x-\bar{x})&\leq k\\
        \Longleftrightarrow && -\frac{1}{2}(x-\bar{x})^TC^{-1}(x-\bar{x})&\geq -\frac{1}{2}k\\
        \Longleftrightarrow && \exp{\left(-\frac{1}{2}(x-\bar{x})^TC^{-1}(x-\bar{x})\right)}&\geq\exp{\left(-\frac{1}{2}k\right)}\\
        \Longleftrightarrow && \frac{\exp{\left(-\frac{1}{2}(x-\bar{x})^TC^{-1}(x-\bar{x})\right)}}{\sqrt{(2\pi)^q\det C}}&\geq\frac{\exp{\left(-\frac{1}{2}k\right)}}{\sqrt{(2\pi)^q\det C}}\\
        \Longleftrightarrow && p(x)&\geq\xi.
    \end{aligned}
\end{equation}
Hence, the hyper-ellipsoid $Q$ contains all likely shapes, \ie, those with a probability greater than a certain threshold, $\xi$.

\begin{algorithm}
    \caption{Metropolis-Hastings algorithm (symmetric proposal distribution)}\label{alg:mh_algorithm}
    \begin{algorithmic}[1]
        \State Initialize $\alpha_0$; set $x_0=f_1(\alpha_0)$.
        \For{$i=0,1,\dots,m$}
            \State Draw sample $\alpha'$ from $Q(\alpha'\, \vert \,\alpha_i)$; set $x'=f_1(\alpha')$.
            \State Compute acceptance ratio as $$t=\frac{L(x';x'\in I_\epsilon)p(x')}{L(x_i;x_i\in I_\epsilon)p(x_i)}.$$
            \State Accept $\alpha'$ with probability $t$ by drawing a sample $r$ from $\mathcal{U}(0,1)$ and
            $$\alpha_{i+1}=
            \begin{cases}
            \alpha' & \text{if } t>r,\\
            \alpha_i & \text{otherwise.}
            \end{cases}$$
        \EndFor
        \State\Return $\{f_1(\alpha_0),f_1(\alpha_1),\dots,f_1(\alpha_m)\}$
    \end{algorithmic}
\end{algorithm}

\section{Implementation Details}
We use Markov chain Monte Carlo (MCMC) to estimate the posterior distribution $p(x \, \vert \, x\in I_\epsilon)$ for $x\in Q_1$ (and $Q_2$, respectively) as stated in Eq. (\ref{eq:post}) of the main paper.
The Markov chain is built by means of the Metropolis-Hastings (MH) algorithm, summarized in Algorithm \ref{alg:mh_algorithm}.
The MH algorithms requires a proposal distribution $Q(\alpha' \, \vert \, \alpha)$, conditioned on the current state $\alpha\in\mathbb{R}^{q_1}$.
We use a random walk mixture proposal of the form:
\begin{equation}
    Q(\alpha'\, \vert \,\alpha)=\sum_{i=1}^n c_i Q_i(\alpha'\, \vert \,\alpha)\quad\text{with}\quad\sum_{i=1}^n c_i=1.
\end{equation}
In our specific implementation, we set $n=4$ and defined:
\begin{equation}
    \begin{split}
        Q_1(\alpha'\, \vert \,\alpha)&=\mathcal{N}(\alpha, 0.2),\quad c_1=0.1,\\
        Q_2(\alpha'\, \vert \,\alpha)&=\mathcal{N}(\alpha, 0.1),\quad c_2=0.5,\\
        Q_3(\alpha'\, \vert \,\alpha)&=\mathcal{N}(\alpha, 0.025),\quad c_3=0.2,\\
        Q_4(\alpha'\, \vert \,\alpha)&=\mathcal{N}(\Vert\alpha\Vert, 0.2),\quad c_4=0.2.
    \end{split}
\end{equation}
This proposal distribution was originally presented in \cite{Schoenborn17} and we left it unchanged.
Note that proposal is \textit{symmetric}, \ie, $ Q(\alpha'\, \vert \,\alpha)= Q(\alpha\, \vert \,\alpha')$, since all mixture components are Gaussian.

Having the proposal distribution in mind, the MH algorithm proceeds as follows: In every iteration, a new sample $\alpha'$ is proposed based only on the previous sample, $\alpha_i$.
The proposed sample is then either accepted or rejected with probability $t$, where $t$ is the so-called acceptance ratio.
After a sufficient number of iterations, $m$, the MH algorithm returns a set of \textit{accepted} samples from the desired posterior distribution, $p(x \, \vert \, x\in I_\epsilon)$.

\section{Evaluation Metrics}
We now provide some additional information on the evaluation metrics used for quantitative analysis as briefly explained in Section \ref{subsec:quant_analysis} of the main paper.

\textbf{Grassmann distance.}
The Grassmann distance is the natural distance between two \textit{linear} subspaces embedded in $\mathbb{R}^n$ (the set of all $k$-dimensional linear subspaces is called the Grassmannian, usually denoted as Gr($k,n$)).
Given orthonormal bases $A,B\in\mathbb{R}^{k\times n}$ for two subspaces from Gr($k,n$), the Grassmann distance can be calculated by means of the principal angles $\{\theta_1,\theta_2,\dots,\theta_k\}$ between the two subspaces.
With slight abuse of notation but for the sake of brevity, we refer to the Grassmann distance as $d_G(A,B)=\Vert(\theta_1,\theta_2,\dots,\theta_k)\Vert_2$.

Since SSMs span \textit{affine} subspaces, the Grassmann distance as presented previously can not directly be applied to measure distances between subspaces spanned by linear shape models.
Fortunately, as shown in \cite{LekHeng20} (Theorem 7), the Grassmann distance can be easily extended to affine subspaces as we briefly explain next.
The key idea is to embed the affine subspace into a linear subspace by adding one dimension.
Given two affine subspaces represented by orthonormal bases $A,B\in\mathbb{R}^{k\times n}$ and displacement vectors $b,c\in\mathbb{R}^n$, their Stiefel coordinates, $Y_1,Y_2\in\mathbb{R}^{(k+1)\times n}$ are given by
\begin{equation}
    Y_1=\begin{pmatrix} A & b_0/\sqrt{1+\Vert b_0\Vert^2} \\ 0 & 1/\sqrt{1+\Vert b_0\Vert^2}
    \end{pmatrix}
\end{equation}
and
\begin{equation}
    Y_2=\begin{pmatrix} B & c_0/\sqrt{1+\Vert c_0\Vert^2} \\ 0 & 1/\sqrt{1+\Vert c_0\Vert^2}
    \end{pmatrix},
\end{equation}
where $b_0$ and $c_0$ are unit vectors orthogonal to the columns of $A$ and $B$, respectively. 
We compute them by 
\begin{equation}
    b'=b-\sum_{j=1}^k (a_j\cdot b) a_j,\quad b_0=\frac{b'}{\Vert b'\Vert_2},
\end{equation}
and analogously for $c_0$.
Here, $a_j\in\mathbb{R}^n$, $j=1,\dots,k$, denote the columns of $A$.
Finally, the affine Grassmann distance is computed by applying Singular Value Decomposition to $Y_1^TY_2$, yielding the $k+1$ principal angles between the respective affine subspaces.
Taking the Euclidean norm of those angles leads to the affine Grassmann distance.

\begin{figure}
    \centering
    \includegraphics[width=\linewidth]{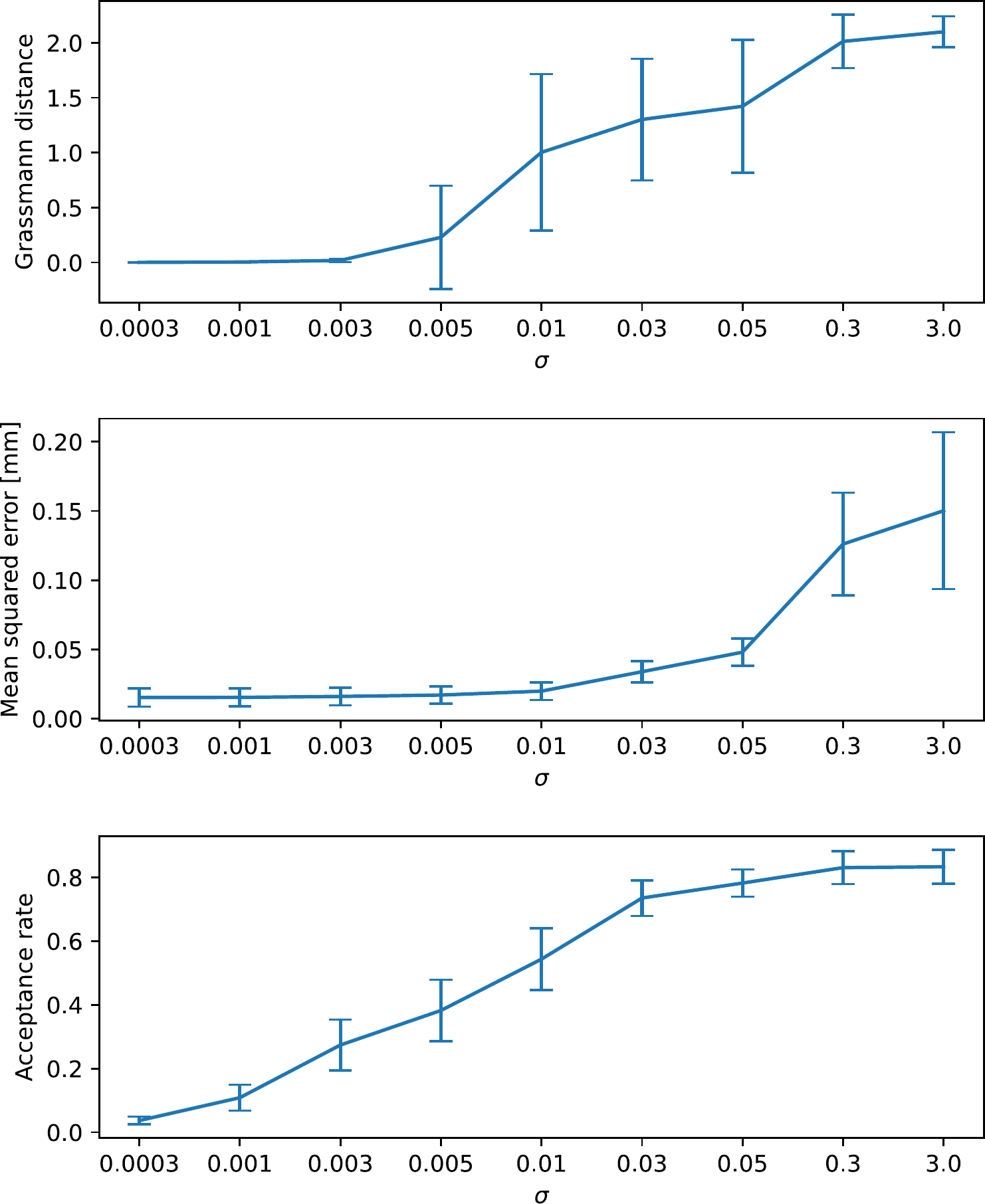}
    \caption{Results of the ablation study for different values of $\sigma$, averaged over all six star models (described in the main paper). Top: Grassmann distance between estimated intersection space and ground-truth intersection; center: mean squared error between MCMC samples and their corresponding projections into the other model (averaged over all posterior samples); bottom: acceptance rates during MCMC sampling.}
    \label{fig:star_ablation}
\end{figure}

\begin{figure*}
    \centering
    \includegraphics[width=\linewidth]{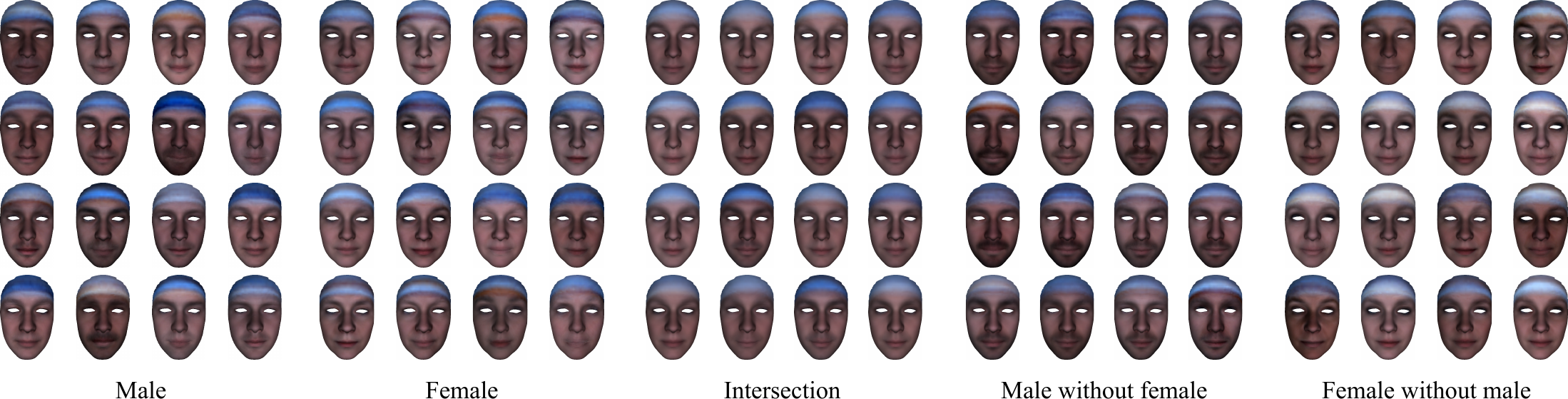}
    \caption{Random samples from the male and female color model of the LYHM~\cite{Dai2019} data set (1st and 2nd block) as well as samples of the computed intersection space (3rd block), and respective differences (male without female, and female without male, 4th and 5th block). Similar to shape (see Figure \ref{fig:male_female} of the main paper), we see stronger male and female dominance in the differences and neutral gender in the intersection, especially visible in the beard region. All color samples are visualized on the mean face.}
    \label{fig:male_female_color}
\end{figure*}

\textbf{Reconstruction error.}
Following the main paper, to evaluate the quality of the computed differences, we exploit the fact that we have samples $\{x_1,x_2,\dots,x_r\}$ from the ground-truth difference, $D_{12}$ (or $D_{21}$).
Denote the MCMC samples from the estimated difference $\hat{D}_{12}$ as $\{\hat{x}_1,\hat{x}_2,\dots,\hat{x}_r\}$.
We then evaluate the quality of $\hat{D}_{12}$ by quantifying whether or not samples $\hat{x}_j$ can be as badly represented in the ground-truth intersection, $I$, as samples $x_j$ from the ground-truth difference.
Here, we expect high errors since shapes belong to the difference if they can \textit{not} be represented in the intersection.
Formally, we calculate the reconstruction errors
\begin{equation}
    d_R(D_{12},I)=\frac{1}{r}\sum_{j=1}^r d(\proj_{M_I}(x_i),x_i)
\end{equation}
and 
\begin{equation}
    d_R(\hat{D}_{12},I)=\frac{1}{r}\sum_{j=1}^r d(\proj_{M_I}(\hat{x}_i),\hat{x}_i)
\end{equation}
by projecting samples onto the subspace $M_I$ spanned by $I$ and evaluating its distance using the mean squared error (MSE; see Eq. (\ref{eq:def_mse}) of the main paper).

\section{Ablation Study}
We identified the variance involved in the distance likelihood, $\sigma^2$, as the most important parameter of our algorithm (see Eq. (\ref{eq:dist_likelihood}) of the main paper). 
To study its effects, in addition to the quantitative evaluation presented in the main text, we also provide an ablation study for different values of $\sigma$.
The ablation is carried out on the star data set as described in Section \ref{subsubsec:synthetic} of the main paper (see also Table \ref{tab:star_model} of the main paper).
Moreover, we only investigate the estimation of ground-truth intersection spaces as our method shows similar behavior for differences.

The results can be found in Figure \ref{fig:star_ablation} (top row), averaged over all six star models.
As seen, setting $\sigma$ too high leads to large Grassmann distances, implying a worse estimation of the ground-truth intersection.
Contrary, the smaller $\sigma$, the better the estimation of the true intersection space.
Interestingly, starting from $\sigma=0.003$, its exact value seems to become less critical as even a decrease of factor 10 does not lead to significant changes.

\section{How to Choose $\sigma$ In Practice?}
Since we usually do not have ground-truth intersections for real-world SSMs, it is natural to wonder how to empirically validate the performance of our method and the chosen $\sigma$ in practice.
To this end, we also report the MSE between an MCMC sample, $x\in Q_1$ (or $x\in Q_2$), and its projection into the other model, $x'$, as well as the acceptance rates during MCMC sampling, see Figure \ref{fig:star_ablation} (center and bottom row).
Note that the MSE is computed between \textit{all} posterior samples and their respective projections; the average thus serves as an empirical estimation for the \textit{mean} $\epsilon$ in Eqs. (\ref{eq:approx_inter}) and (\ref{eq:def_approx_inter}) of the main text.

In terms of MSE, we observe an almost similar behavior as for the Grassmann distances.
Smaller MSEs correspond to lower Grassmann distances.
As a result of this correlation, the distance between $x$ and $x'$ can be used as an indication to determine a suitable value for $\sigma$.
It does not require a ground-truth and can be easily monitored during the run time of MCMC.
To avoid setting $\sigma$ too small (and preventing incorrect estimation of the desired distribution), however, one should ensure that the acceptance rates are between 0.25 and 0.5, see \cite{Schoenborn17}.

In conclusion, although a suitable value for $\sigma$ might not be trivial to determine in practice, it can be well chosen by carefully inspecting the MSE between posterior samples and their corresponding projections as well as acceptance rates during MCMC sampling.
Both quantities can be easily computed without the necessity of ground-truth intersections or differences.

\begin{figure}
    \centering
    \includegraphics[width=\linewidth]{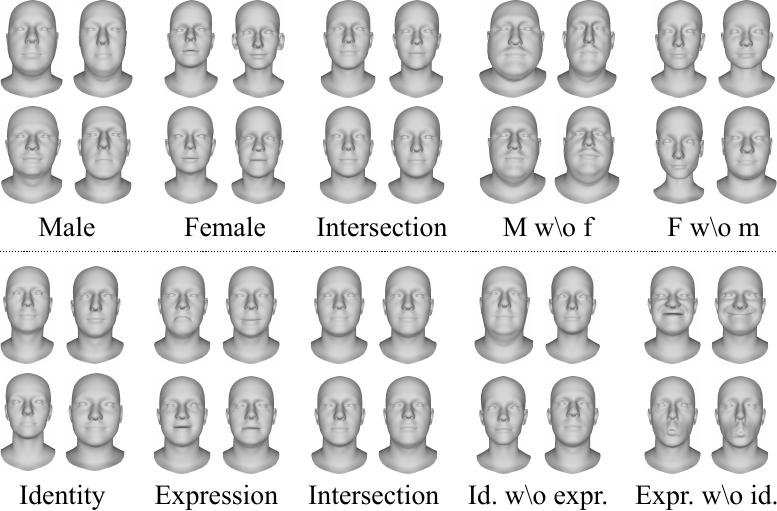}
    \caption{Random samples from the male and female models (top) and identity and expression models (bottom) of FLAME (1st and 2nd block). Also shown are samples of the computed intersection model (3rd block) as well as from the difference between male and female (top) and identity and expression models (bottom; 4th and 5th block). We observe very similar results as for LYHM models and BFM 2019~\cite{gerig2018morphable}, see Figures \ref{fig:male_female} and \ref{fig:id_expression_samples} of the main paper.}
    \label{fig:flame_idexpr_malefemale}
\end{figure}

\begin{figure}
    \centering
    \includegraphics[width=\linewidth]{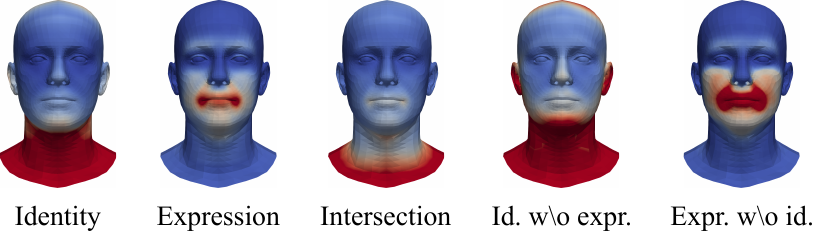}
    \caption{Per-vertex level visualization of similarities and differences between the identity and expression spaces of FLAME. Depicted is the per-vertex variance of identity and expression models as well as the posterior variance for the computed intersection model and differences (red high variance, blue low). 
    We observe a similar behavior as for BFM 2019, see Figure \ref{fig:identity_expression} of the main paper.}
    \label{fig:flame_postvar}
\end{figure}
\section{Additional Qualitative Results}
In this section, we provide results for the extension to color experiment (Section \ref{subsec:extension_color} of the main paper) as well as additional qualitative results on the FLAME \cite{Li17} model.

\textbf{Extension to color.}
The results are shown in Figure~\ref{fig:male_female_color}.
We observe very similar behavior for the color as we have seen for shape, see Figure \ref{fig:male_female} of the main paper. 
Male features are exaggerated in faces drawn from the difference of male and female, and samples from the difference of female and male appear more feminine than faces from the original female model. 
Similar to the effect in shape, we also perceive neutral textures in the intersection space.

\textbf{Results on FLAME.} 
We show exemplary results on FLAME in Figures \ref{fig:flame_idexpr_malefemale} and \ref{fig:flame_postvar}.
They are very similar to the findings presented in the main paper, please refer to Figures \ref{fig:male_female}--\ref{fig:identity_expression} of the main paper.

\end{document}